# Subjective and Objective Evaluation of English to Urdu Machine Translation

Vaishali Gupta[#1], Nisheeth Joshi[#2], Iti Mathur[#3]

[#]Apaji Institute, Banasthali University, Rajasthan, India

```
1vaishali.gupta77@gmail.com,
2nisheeth.joshi@rediffmail.com,
3mathur_iti@rediffmail.com
```

*Abstract*- **Machine translation is research based area where evaluation is very important phenomenon for checking the quality of MT output. The work is based on the evaluation of English to Urdu Machine translation. In this research work we have evaluated the translation quality of Urdu language which has been translated by using different Machine Translation systems like Google, Babylon and Ijunoon. The evaluation process is done by using two approaches – Human evaluation and Automatic evaluation. We have worked for both the approaches where in human evaluation emphasis is given to scales and parameters while in automatic evaluation emphasis is given to some automatic metric such as BLEU, GTM, METEOR and ATEC.**

*Keywords*- **BLEU, GTM, METEOR and ATEC.**

## I. INTRODUCTION

Evaluation plays a major role in the field of Natural Language Processing. Evaluation is necessary for development of Machine Translators. To overcome the language barrier problem, researchers have tried to design the MT systems. To ascertain the performance of MT system, we employ evaluation process. So that we may get precise report of MT development process. Evaluation depends on the subject matter, applied methodology or the application of its results. In general, evaluation can be understood as judgment on the value of a public intervention with reference to defined criteria of this judgment. In this paper, we are doing sentence level evaluation. The goal of this paper is three fold: Human Evaluation, Automatic Evaluation and Correlation between Human and Automatic Evaluation.

In Human Evaluation, to check the quality of MT output, human expert is required who knows that particular language as human expert is best evaluator to judge the quality of MT output and also he provides the feedback for development of MT system. There is some drawback with human evaluation as, it is time consuming, costly and also it gives subjective judgment score. So it becomes difficult to analyze anything for a particular MT output.

In Automatic Evaluation, we are using automated metrics like BLEU, GTM, METEOR and ATEC to check the quality of MT output. These automated metrics are more beneficial than human evaluator because these metrics provide quick evaluation score and evaluate large data set in lesser time. Automated metrics are repeatable i.e. when we give the same input in particular metrics can give the same results. So we can say that results obtained through human evaluation varies from human to human and for the same data set, it is not possible to get same score every time. Despite of these many features, automated metrics is not sufficient for measuring the quality of MT output. Eventually we need a human judgment. In this way, we can say that for the development of MT systems both of these approaches are important.

Finally we correlate both of these approaches and contemplate that which automatic metric convey proximate result to human evaluation score. That means the metrics which gives the close result with human evaluation score which are highly correlated with human judgments.

## II. RELATED WORK

In this area many researchers have analyzed the quality of MT output. They have also proposed some approaches for evaluation purpose. Initially these MT outputs are evaluated by human expert. Since humans require a lot of time and money, therefore the researchers developed automated metrics for automatic evaluation. Snover et al [1] proposed a study of translation edit rate with targeted human annotation. In this study, authors described new approach for checking the quality of MT output. Joshi et al [2] proposed human and automatic evaluation of English to Hindi Machine Translation. In this paper, the authors described some scale based adequacy and fluency measures for human evaluation. They have designed METEOR for Hindi to calculate automatic score and then give correlation between human and automatic evaluation. Papineni [3] proposed a BLEU metric for automatic evaluation of MT. BLEU metric is quick, inexpensive and language independent and also correlates highly with human evaluation. Turian et al [5] proposed a General Text Matcher. In this paper, the authors described evaluation technique like precision, recall and F-measure. They showed that F-measure is highly correlated with human judgments. Lavie and Agarwal [6] proposed METEOR metric which is an automatic metric for MT evaluation with high levels of correlation with human judgments. In this paper, the authors described an approach for the implementation of METEOR for Spanish, French and German language. Wong & Kit [7] proposed ATEC metric for automatic evaluation, which basically depends on two essential features: Unigram word choice and word position. Computation of this metric is based on Unigram F-measure. Coughlin [8] presented a paper on correlating automated and human assessment of MT quality. In this paper, the author described the human evaluation and automatic evaluation and then correlated human score with BLEU and NIST metrics. Agarwal and Lavie [9] described the correlation between human ranking of MT and evaluation metric METEOR, M-Bleu and M-TER. Here the author showed improvement in correlation as compared to earlier Metric and also described M-Bleu and M-TER metrics.

## III. HUMAN EVALUATION

Human evaluation is done by human annotator. Fig. 1 shows the process of human evaluation is shown below:

**Fig. 1.** Human Evaluation Process

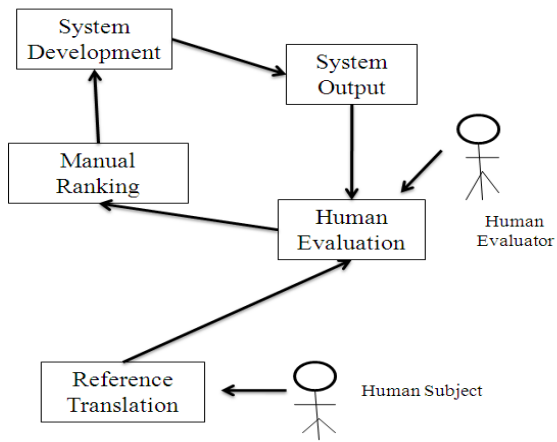

For human evaluation, primarily we assembled 1000 sentences from health and tourism domain to make the corpus. In all we took 1000 sentences which were divided into 10 docs of 100 sentence each. After that we registered Urdu MT outputs for each sentence of the corpus using MT engines such as Google, Babylon and Ijunoon. Then we evaluated 3000 (1000×3) MT outputs manually. Here human evaluation was based on 5 scales and 10 parameters [2]. These scales and parameters are as follows:

*A. Scale*

1. Not Acceptable (0)
2. Partially Acceptable (1)
3. Acceptable (2)
4. Perfect (3)
5. Ideal (4)

*B. Parameter:*

1. Translation of Gender and Number of the Noun/s.
2. Translation of tense in the source sentence.
3. Translation of Voice in the source sentence.
4. Identification of the Proper Nouns.
5. Use of Adjectives and Adverbs corresponding to the nouns and verbs in the source sentence.
6. Selection of proper words / synonyms.
7. The sequence of Noun, Helping Verb and Verb in the translation.
8. Use of Punctuation signs in the translation.
9. Maintaining the stress on the significant part in the source sentence in the translation.
10. Maintaining the semantics of the source sentence in the translation.

To explain the human evaluation, let us take an example:

*Source Sentence:*

Taj mahal is in india , made by shahjahan

*Target Sentence:*

*Google:*  تاج محل شاہجہاں کی طرف سے بنایا، بھارت میں ہے

(*taajmahal shahjahan ki taraf se bnaya bharat men hai*)

*Babylon:*  بھارت میں تاج محل, شاہ جہان نے

(*taajmahal men bharat ki jaanab se shahjahan*)

*Ijunoon:*  تاج ماحل ہے میں انڈیا ، بنا بزریعہ شاہ جہان

(*taajmahal hai mein andia, bna bazariyah shah jahan*)

These MT outputs are evaluated on the basis of scale and parameters. These scores are as follows:

TABLE I
Human Evaluation of MT Output

| Parameter | Google | Babylon | Ijunoon |
|---|---|---|---|
| 1 | 2 | 0 | 1 |
| 2 | 3 | 1 | 2 |
| 3 | 2 | 1 | 1 |
| 4 | 2 | 2 | 2 |
| 5 | 2 | 0 | 0 |
| 6 | 2 | 1 | 2 |
| 7 | 1 | 0 | 1 |
| 8 | 2 | 1 | 1 |
| 9 | 2 | 1 | 1 |
| 10 | 2 | 1 | 1 |
| Average | 2.0 | 0.8 | 1.2 |

From the above table, we observe that Google provides us the maximum score output for the input that we have given. On a scale of 0-4 it gives an accuracy of 2. These scale points are then converted into percentage. '2' scale point means the score is average that means 50% accurate. Similarly Babylon gives us the score as 0.8 which lies between 0 to 4 scales. After calculating the percentage, it gives 20% accuracy. Finally the output from Ijunoon is obtained as 1.2 score which means 30% accurate.

IV. AUTOMATIC EVALUATION

In automatic evaluation, we introduced Similarity based metrics viz. BLEU, GTM, METEOR and ATEC. These metrics are highly correlated with human judgments and are used for evaluation of various language pair. In this paper, we are describing these metrics only for English-Urdu language pair because we have used Urdu stem matching and Urdu synonyms for the development of these metrics. Therefore these metrics give evaluation score after mapping the Urdu MT output and Urdu reference output. The brief introduction of these metrics is as follows. Papineni [3] introduced the BLEU metric that support for n-gram calculation. BLEU metric is improved by NIST metrics, proposed by Doddington [4]. Both BLEU/NIST metrics having some drawback that it is weakly correlated with human judgment of translation quality. Then Turian et al [5] proposed a GTM metric that is based on F-measure. F-measure is highly correlated with human judgment rather BLEU/NIST. Then these metrics are improved by METEOR, proposed by Lavie and Agarwal [6]. METEOR metric improve the translation quality of MT output. As METEOR is not only used for word to word matching between MT output and reference output. It also uses stem and synonym matching. Hence it gives good correlation between human and automatic evaluation. Later, Wong and Kit [7] proposed ATEC metric that is used for calculating score of MT output on the basis of word choice and word order phenomenon. In automatic evaluation some components are used for the evaluation purpose. These are as follows:

- Reference Output: Translation of source sentence by human expert.
- MT output: It gives by MT engines.
- Precision (P): Matched words with respect to MT output.
- Recall (R): Matched words with respect to reference output.
- F-measure: $\frac{2PR}{P+R}$

A. *BLEU-(Bilingual Evaluation Understudy):*

In 2000, Papineni [3] proposed BLEU metric in IBM. It is based on n-gram precision measure and is totally depends on the geometric average of n-gram matching between MT output and Reference output. Formula for brevity penalty (BP) and BLEU is as follows:

$$BP = \min(1, \frac{output-length}{reference-length}) \quad (1)$$

$$BLEU = BP * (\prod_{i=1}^{n} precision_i)^{1/n} \quad (2)$$

B. *GTM-(General Text Matcher):*

Turian et al established GTM metric which was based on the idea of Melamed et al, 2003. Through the sharing of matched words between MT output and reference output, we provide the evaluation score for MT output. Unlike BLEU, it is not only based on precision and recall. Score calculation of GTM is based on harmonic mean of precision and recall, also known as F-measure.

$$F\text{-measure}: \frac{2PR}{P+R} \quad (3)$$

C. *METEOR-(Metric for Evaluation of Translation with Explicit Ordering):*

In 2004, Meteor was proposed by Lavie et al. It was developed explicitly for higher correlation with human judgment to improve the quality of MT engine at segment level. In this paper, we have implemented Meteor metric for the evaluation of English to Urdu Machine Translation. METEOR compute a score, based on word to word matching between MT output and reference output. Initially Meteor creates a word to word alignment between two strings. This alignment is increased by Urdu stem matching along with Urdu synonyms matching. This type of matching is called unigram matching, then through this alignment meteor compute an evaluation score between MT output and reference output. Now, we calculate total number of unigram in MT output, reference output and matched unigram in both the string. Then we calculate unigram precision and recall, by using this parameterized harmonic mean is computed as:

$$\text{F-mean} = \frac{2PR}{P+R} \quad (4)$$

After unigram matching we proceed for chunks matching. Chunks means adjacent set of words. So if we find same adjacent set of words in both the string then we can count it as a chunk. For a given alignment, Meteor also computes the penalty. For calculating the penalty, it uses the number of chunks (*ch*) and number of matched unigram (*m*) as shown below:

$$Penalty = \frac{Chunks\ (ch)}{Matched\ Unigram\ (m)} \quad (5)$$

Finally Meteor score is calculated by this formula:

$$Score = (1 - penalty) * F - mean \quad (6)$$

D. ATEC:

Wong and Kit introduced ATEC metric. ATEC metric uses two essential features that is word choice and word position for evaluating the quality of MT output. Computation of this metric is based on unigram F-measure, which describe word to word matching between MT output and reference output and also describe the average difference of relative position of matched word. Here ATEC metric is used for Evaluation of English-Urdu language pair as we are using Urdu stem and Urdu synonym at the time of implementation. Next we provide detailed description of word choice and word position.

*1) Unigram-based measure of word choice:*

We measure the word choice of a translation by unigram matching rate, which can be represented by the standard measures of precision (P) and recall (R). Here number of matched unigram (M) between a machine translation (m) and reference translation (r) and length of machine output (|m|) and length of reference translation (|r|) is used for calculating precision and recall.

$$Precision(P) = \frac{M(m,r)}{|m|} \quad (7)$$
$$Recall(R) = \frac{M(m,r)}{|r|} \quad (8)$$

We are also calculating F-measure (F) to know the less or more words in machine output than its reference output. F-measure is the average of precision and recall.

$$F - measure(F) = \frac{2PR}{P+R} \quad (9)$$

*Example-1 (i) for word choice:*

Here we are maximizing the unigram matches between a machine translation output and reference output. As in Example-1 MT output matches 9 words (in underline) with reference output.

*Reference output:* Bhopal is a Lake City and capital of Madhya Pradesh.
*MT output:* Bhopal is the capital of Madhya Pradesh and also called Lake City.

*2) Penalty of Word Position difference:*

In this, we are measuring the position of words for matching the Machine Translation output and reference output. Generally we think that every word has its appropriate position in a sentence to contribute for the meaning of a particular sentence. In example 2(a), MT output-1 has a different meaning from the reference output although they share the same words. MT output-2 shares many consecutive words with the reference output but it is grammatically incorrect. MT output-3 matches the least words with the reference output, but it has the closest meaning to it.

*Example 2(a):*

*Reference output:* manager works with our employee.

*MT output-1:* employee works with our manager.
*MT output-2:* works employee with our manager.
*MT output-3:* manager fairly works with our employee.

For counting their variances in position of word order, we first assign an absolute position to each of the words of both MT outputs and references output. The absolute positions are then converted to relative positions by dividing them to the lengths of MT output or reference output in order to normalize the length difference of each sentence, as shown in example 2(b).

*Example 2(b):*

*Reference:* manager works with our employee.
Absolute position: 1  2  3  4  5
Relative position: 0.2  0.4  0.6  0.8  1

*MT-1:* employee works with our manager.
Absolute position: 1  2  3  4  5
Relative position: 0.2  0.4  0.6  0.8  1

*MT -2:* works employee with our manager.
Absolute position: 1  2  3  4  5
Relative position: 0.2  0.4  0.6  0.8  1

*MT-3:* manager fairly works with our employee.
Abs. pos.: 1  2  3  4  5  6
Rel. pos.: 0.17  0.33  0.5  0.67  0.83  1

For all MT output, each word is aligned to their corresponding words in the reference output. After this alignment process, position difference is calculated by taking the difference between MT string and reference string. Then sum of this position difference is divided by the length of MT string.

*Example 2(c):*

*Reference:* manager works with our employee.
0.2  0.4  0.6  0.8  1
*MT -1:* employee works with our manager.
0.2  0.4  0.6  0.8  1
Position Difference= (0.8+0+0+0+0.8)/5 = 0.32

*Reference:* manager works with our employee.
0.2  0.4  0.6  0.8  1
*MT-2:* works employee with our manager.
0.2  0.4  0.6  0.8  1
Position Difference= (0.8+0.2+0+0+0.8)/5 = 0.36

*Reference:* manager works with our employee.
0.2  0.4  0.6  0.8  1
*MT-3:* manager fairly works with our employee.
0.17  0.33  0.5  0.67  0.83  1
Position Difference= (0.3+0.1+0.07+0.03+0)/6= 0.083

After calculating the position difference between a one or more MT outputs and references output, it is then converted to a penalty rate for the MT output. According to empirical experiment of Wong and Kit [8], the word position difference has to be multiplied by a coefficient 4 for the highest correlation with human judgment.

$$Penalty = 1 - (Posdiff * 4) \qquad (10)$$

If the word position difference of a MT output is greater than 0.25, as in *MT-1* and *MT-2* of example 2(c), the penalty will be negative. In this case the penalty will be approximate to 0. Finally score of ATEC metric is calculated by this formula:

$$ATEC = F - measure * Penalty \qquad (11)$$

## V. DISCUSSION AND RESULTS

For showing the results of human evaluation and automatic evaluation approaches, we are taking a common example as shown below:

*Source:* Taj mahal is in india , made by shahjahan.

*Target:*

*Ref.:* تاج محل بھارت میں ہے ، جو کہ شاہجہاں نے بنوایا تھا

(*taaj mahal bharat mein hai , jo ki shahjahan ne bnvaya tha* )

*Google:* تاج محل شاہجہاں کی طرف سے بنایا، بھارت میں ہے

 (*taaj mahal shahjahan ki taraf se bnaya bharat men hai*)

*Babylon:* بھارت میں تاج محل, شاہ جہان نے

 (*taaj mahal men bharat ki jaanab se shahjahan*)

*Ijunoon:* تاج ماحل ہے میں انڈیا ، بنا بزریعہ شاہ جہان

 (*taaj mahal hai mein andia, bna bazariyah shah jahan*)

For this example, evaluation score is calculated by human expert and automated metrics that is shown in below table:

TABLE II
Score of human and automatic evaluation

|  | BLEU | GTM | Meteor | ATEC | Human |
|---|---|---|---|---|---|
| **Google** | 0.63 | 0.67 | 0.58 | 0.21 | 0.50 |
| **Babylon** | 0.33 | 0.36 | 0.13 | 0.48 | 0.20 |
| **Ijunoon** | 0.34 | 0.36 | 0.34 | 0.36 | 0.30 |

## VI. CORRELATION

We calculate correlation between human and automatic evaluation by using Pearson's rank correlation formula as follows:

$$r = \frac{\sum dxdy}{\sqrt{\sum dx^2 dy^2}} \quad (12)$$

Where -

$$\sum dx^2 = (\sum x^2 (\sum x)^2)/n \quad (13)$$

$$\sum dy^2 = (\sum y^2 (\sum y)^2)/n \quad (14)$$

$$\sum dxdy = \sum xy - \frac{\sum x \sum y}{n} \quad (15)$$

Now we have described two level of correlation:

### A. Sentence Level

In the sentence level, score is calculated by the metric for particular Machine Translated sentence and then correlated with human judgment. Sentence level correlation is obtained for above example as follows:

TABLE III
Correlation score for sentence level

|  | BLEU | GTM | Meteor | ATEC |
|---|---|---|---|---|
| **Google** | 0.0967 | 0.1093 | 0.1232 | 0.1134 |
| **Babylon** | 0.0725 | 0.0916 | 0.1142 | 0.0996 |
| **Ijunoon** | 0.0913 | 0.1012 | 0.0987 | 0.1071 |

### B. Corpus Level

In this corpus level, aggregate score are calculated over the set of sentences of both human judgment and metric judgment. In this paper, we have taken 1000 sentence of corpus and we get 3000 MT output by three MT engine. Then we calculate aggregate score over the generated machine translated sentences by metric judgment and human judgment. Corpus level correlation score is provide in below table:

TABLE IIIV
Correlation score for corpus level

|  | BLEU | GTM | Meteor | ATEC |
|---|---|---|---|---|
| **Google** | 0.0918 | 0.1012 | 0.1312 | 0.1098 |
| **Babylon** | 0.0725 | 0.0876 | 0.1413 | 0.0886 |
| **Ijunoon** | 0.0911 | 0.0912 | 0.0915 | 0.0901 |

## VII. CONCLUSION

In this paper, we have demonstrated various evaluation approaches for measuring the quality of MT output. Firstly, we described the scale and parameter based human evaluation and then some described some automatic metrics. Among these automatic metrics METEOR and ATEC metric are implemented on language specific instances. All the metrics have been used for English to Urdu machine translated output. We also described the correlation between human judgment and automated metrics i.e. the metrics which gives closest score to human judgment, that metric is considered to be the best. Through the correlation section, we conclude that METEOR metric is highly correlated with human judgment followed by ATEC.